\documentclass[11pt,a4paper]{article}
\usepackage[hyperref]{acl2020}
\usepackage{times}
\usepackage{footnote}
\usepackage{latexsym}

\usepackage{scrextend}
\usepackage{latexsym}
\usepackage{amsmath}

\newcommand{\tf}[1]{\textbf{#1}}
\usepackage{amssymb,amsmath,amsthm,enumitem}
\usepackage{comment}

\usepackage{url}
\usepackage{graphics}
\usepackage{tablefootnote}

\usepackage{times}
\usepackage{graphicx}
\usepackage{amsmath,amssymb,enumerate,bbm,color,alltt}
\usepackage{graphicx}
\usepackage{float}
\usepackage{stfloats}
\usepackage{subfig}
\usepackage{longtable}
\usepackage{tabularx}
\usepackage{multirow}
\usepackage{booktabs}
\usepackage{subfig}
\usepackage{url}

\usepackage{hyperref}
\usepackage{url}
\usepackage{tabularx}
\newcolumntype{C}[1]{>{\centering\arraybackslash}p{#1}}
\usepackage{wrapfig,lipsum,booktabs}
\usepackage{amsmath}
\usepackage{amssymb}
\usepackage{subfig}
\usepackage{adjustbox} 
\usepackage{caption}

\usepackage{microtype}

\aclfinalcopy 

\title{A Simple but Tough-to-Beat Data Augmentation Approach for \\ 
    Natural Language Understanding and Generation}

\author{Dinghan Shen$^{\mathbf{1}}$, ~Mingzhi Zheng$^{\mathbf{1}}$, ~Yelong Shen$^{\mathbf{1}}$, ~Yanru Qu$^{\mathbf{2}}$, ~Weizhu Chen$^{\mathbf{1}}$ \\
	\smallskip 
	$^{\mathbf{1}}$ Microsoft Dynamics 365 AI~~~~
	$^{\mathbf{2}}$ University of Illinois at Urbana-Champaign~~~ \\
	\smallskip
	$^{\mathbf{1}}$ {\tt \{dishen, mizheng, yeshe, wzchen\}@microsoft.com} \\
	$^{\mathbf{2}}$ {\tt \{yanruqu2\}@illinois.edu} \\
  }

\begin{document}
\maketitle

\begin{abstract}
Adversarial training has been shown effective at endowing the learned representations with stronger generalization ability. However, it typically requires expensive computation to determine the direction of the injected perturbations. In this paper, we introduce a set of simple yet efficient data augmentation strategies dubbed \emph{cutoff}, where part of the information within an input sentence is erased to yield its restricted views (during the fine-tuning stage). 
Notably, this process relies merely on stochastic sampling and thus adds little computational overhead.
A Jensen-Shannon Divergence consistency loss is further utilized to incorporate these augmented samples into the training objective in a principled manner.
To verify the effectiveness of the proposed strategies, we apply \emph{cutoff} to both natural language understanding and generation problems. 
On the GLUE benchmark, it is demonstrated that cutoff, in spite of its simplicity, performs on par or better than several competitive adversarial-based approaches.
We further extend \emph{cutoff} to machine translation and observe significant gains in BLEU scores (based upon the Transformer Base model). Moreover, \emph{cutoff} consistently outperforms adversarial training and achieves state-of-the-art results on the IWSLT2014 German-English dataset.
The source code can be obtained from: \url{https://github.com/dinghanshen/Cutoff}.
\end{abstract}

\section{Introduction} 
Large-scale language models (LMs) pre-trained with massive unlabeled text corpora, in a self-supervised manner, has brought impressive performance gains across a wide range of natural language processing tasks \cite{devlin2018bert, liu2019roberta, yang2019xlnet, joshi2019spanbert, sun2019ernie, clark2019electra, lewis2019bart, bao2020unilmv2, he2020deberta}. Significant research efforts have focused on exploring various pre-training recipes to yield more informative LMs. However, given the imbalanced nature between the huge number of model parameters and limited task-specific data, how to leverage and unlock the knowledge from large-scale LMs (during the fine-tuning stage) remains a challenging issue. It has been observed that the representations from pre-trained models, after being fine-tuned on specific downstream tasks, tend to degrade and become less generalizable \cite{zhu2019freelb, jiang2019smart, aghajanyan2020better}.

To alleviate this issue, adversarial training objectives have been proposed to regularize the learned representations during the fine-tuning stage \cite{zhu2019freelb, liu2020adversarial, jiang2019smart}. Specifically, label-preserving perturbations are performed on the word embedding layer, and the model is encouraged to make consistent predictions regardless of these noises. Although the model's robustness can be improved with these perturbed examples, adversarial-based methods typically require additional backward passes to decide the direction of the inject perturbations. 
As a result, these methods give rise to significantly more computational and memory overhead (relative to standard SGD training).

In this paper, we introduce a set of simple yet efficient data augmentation strategies. They are inspired by the consensus principle in multi-view learning \cite{Blum1998CombiningLA, Xu2013ASO, Clark2018SemiSupervisedSM}, which states that maximizing the agreement/consensus between two different views of data can lead to lower error rate.
Specifically, we propose to erase/remove part of the information within a training instance to produce multiple perturbed samples. 
To ensure that the model cannot utilize the information from the removed input at all, the erasing process happens at the input embeddings layer.
In contrast to Dropout, which converts individual elements within the word embedding matrix to $0$, we propose to erase the vectors along each dimension entirely.
As a result, either multiple tokens or embedding dimensions are converted to vectors of all zeros, yielding partial views of the input matrix in a structured manner.  

To make the augmented samples more challenging, inspired by \cite{joshi2019spanbert}, we further introduce an approach to derive restricted views by removing a contiguous span within an input sequence.
The model is fine-tuned with the constraint of making consistent predictions on these augmented data (even with partial views of the original input).
Intuitively, the resulting representations tend to have a stronger ability of \emph{fully} abstracting various semantic features from a sentence, since the model can not merely utilize the most salient ones (which may not be available in partial views) to make the corresponding predictions.

To capture the intrinsic relationship among these stochastic and diverse augmented examples, we propose a specially-designed consistency regularization objective. Particularly, in addition to the cross-entropy loss typically employed in data augmentation, a Jensen-Shannon Divergence consistency loss is further introduced to match the predictions between different partial views of a given input. One advantage is that this loss is able to naturally maximize the consensus between multiple (more than $2$) views in a more principled and stable manner. 

We evaluate the effectiveness of the proposed data augmentation strategies on a wide range of natural language understanding (NLU) tasks from the GLUE benchmark. RoBERTa \cite{liu2019roberta} is employed as the testbed model in our experiments. However, the augmentation methods proposed here can be easily extended to other large-scale pretrained models. 
Despite its simplicity, our method consistently gives rises to significant performance gains.
More importantly, \emph{cutoff} outperforms several competitive adversarial-based approaches, while being much more computationally efficient. We further extend \emph{cutoff} to the text generation scenario and verify it on a machine translation task. The proposed methods greatly outperform adversarial training on both WMT2014 English-to-German and IWSLT2014 German-to-English tasks. In addition, while combining \emph{cutoff} with a Transformer base model, we achieved state-of-the-art test result on the IWSLT2014 German-to-English dataset, with a BLEU score of $37.6$.
\vspace{-0.5mm}
\section{Related Work}
\vspace{-1mm}
\paragraph{Adversarial Training}
Adversarial training was originally proposed to attack neural-network-based models by applying small perturbations to the input \cite{szegedy2013intriguing}.
Thereafter, several adversarial-based approaches, including adversarial examples \cite{goodfellow2014explaining}, PGD \cite{aleks2017deep}, \emph{etc}, have been introduced. It has been demonstrated that these methods can improve the robustness and generalization ability of a model by augmenting the perturbed examples into the original training instances. Recently, adversarial-based approaches emerged as a popular research trend in NLP, which have been successfully applied to a wide variety of NLU tasks, including sentence classification, machine reading comprehension (MRC) and natural language inference (NLI) tasks, \emph{etc}. Despite its success, computational overhead is typically required to calculate the perturbation directions. Several research efforts have been devoted to accelerate adversarial training \cite{shafahi2019adversarial, zhang2019you}. 
However, additional forward-backward passes are still needed for adversarial training. Our proposed cutoff methods are much more computationally efficient from this perspective. 
Besides, the connection  between adversarial training and data-augmentation-based approaches has not previously been well-established. 
Our work bridges this gap by unifying the two types of methods under the consistency training framework. 

\nocite{Chen2020MixTextLI}

\vspace{-2mm}
\paragraph{Multi-view Learning}
The main idea of multi-view learning is to produce distinct subsets (views) of features corresponding to the same data, and the predictions by the model according to different views are repelled to be consistent \cite{Xu2013ASO}. Our approach is slightly different from such algorithms, \emph{e.g.}, co-training \cite{Blum1998CombiningLA} and co-regularization \cite{sindhwani2005co}, in the sense that the multiple views from \emph{cutoff} have certain overlaps, rather than being entirely independent. 

The intuition of our method bears resemblance to cross-view training (CVT) \cite{Clark2018SemiSupervisedSM}, which also proposes to improve sentence representations by encouraging consistent predictions across different views of the input. However, there are several key differences that make our work unique (except that CVT focuses on a  semi-supervised setting, rather than a supervised one as in our case):
\emph{\romannumeral1}) CVT generates partial views on top of latent representations, while \emph{cutoff} operates at the input embedding layer. As a result, our method is more generic and model-agnostic;
\emph{\romannumeral2}) CVT adds an auxiliary prediction module during the training stage, while span cutoff requires no changes to the original model at all;
\emph{\romannumeral3}) we leverage Jensen-Shannon Divergence consistency loss to match the predictions with various views, which maximize their consensus in a more natural and stable manner (also more efficient than the multiple KL divergence terms used in CVT). 
\vspace{-1mm}
\section{Proposed Approach}
\vspace{-1mm}
In this section, we first discuss the motivation behind the \emph{cutoff} data augmentation strategies, which leverage restricted views of a training instance. Then, we propose three simple but effective ways of obtaining partial views and highlight the advantages of each. A novel consistency loss is introduced to naturally integrate multiple cutoff samples into the training framework. Finally, we further extend the \emph{cutoff} approach to the text generation scenario.  

\subsection{Motivation}
In the context of fine-tuning large pre-trained models, our hypothesis is that data augmentation could endow the original data with stronger ability to extract useful semantic information. Let $f$ denote the transformation to convert an input sample $x$ into its augmented examples. An ideal $f$ should be label-preserving, \emph{i.e.}, the label of $f(x)$ should be the same as $x$. Besides, $f(x)$ should also be diverse and different enough from $x$, so that it could help to enrich the empirical observations and thus better cover the data space. 

Different choices of $f$ to introduce slight modifications on the original training instances have been proposed previously, such as adding Gaussian noise, adversarial training \cite{liu2020adversarial, zhu2019freelb} and back-translation \cite{wei2018fast, xie2019unsupervised}. Concretely, adversarial training performs perturbations on the word embedding layer to improve the model robustness. 
However, it takes additional backward passes to estimate the optimal perturbation direction and thus gives rise to additional computational and memory overhead.
As to back translation, it first translates an existing example to another language, and then translate it back to obtain an augmented sample. 
Although effective, the quality of augmented data is usually sensitive to the mistakes made by the initial translation models \cite{chapelle2009semi, Wang2018SwitchOutAE}. Motivated by these observations, we aim to propose data augmentation schemes that are computationally efficient, and yet do not rely on any additional models or external data sources (\emph{e.g.}, paired translation data). 

Taking inspiration from multi-view learning, where the connection between the consensus of predictions on two views and their error rates has been established \cite{dasgupta2002pac}. Suppose we have two views $x_1$ and $x_2$ obtained with the same example, and let $p_1$ and $p_2$ denote a model's prediction on these views, respectively. The following holds true \cite{dasgupta2002pac}:
\begin{align}
\vspace{-4mm}
P(p_1 \neq p_2)  \geq \max \{P_{err}(p_1), P_{err}(p_2) \}
\label{eq:multi_view}
\vspace{-4mm}
\end{align}
The inequality above states that the error rates of the two hypotheses are both upper bounded by the probability of a disagreement between them. In other words, the accuracy of each prediction can be improved by minimizing their disagreement. Thus, encouraging consistent predictions among different views of a sentence, which contains only part of its entire information, could improve the generalization ability of the resulting models and reduce their error rates accordingly. In the next section, we will discuss how these views may be produced in detail. 
\begin{figure*}
	\centering
	\includegraphics[width=0.75\textwidth]{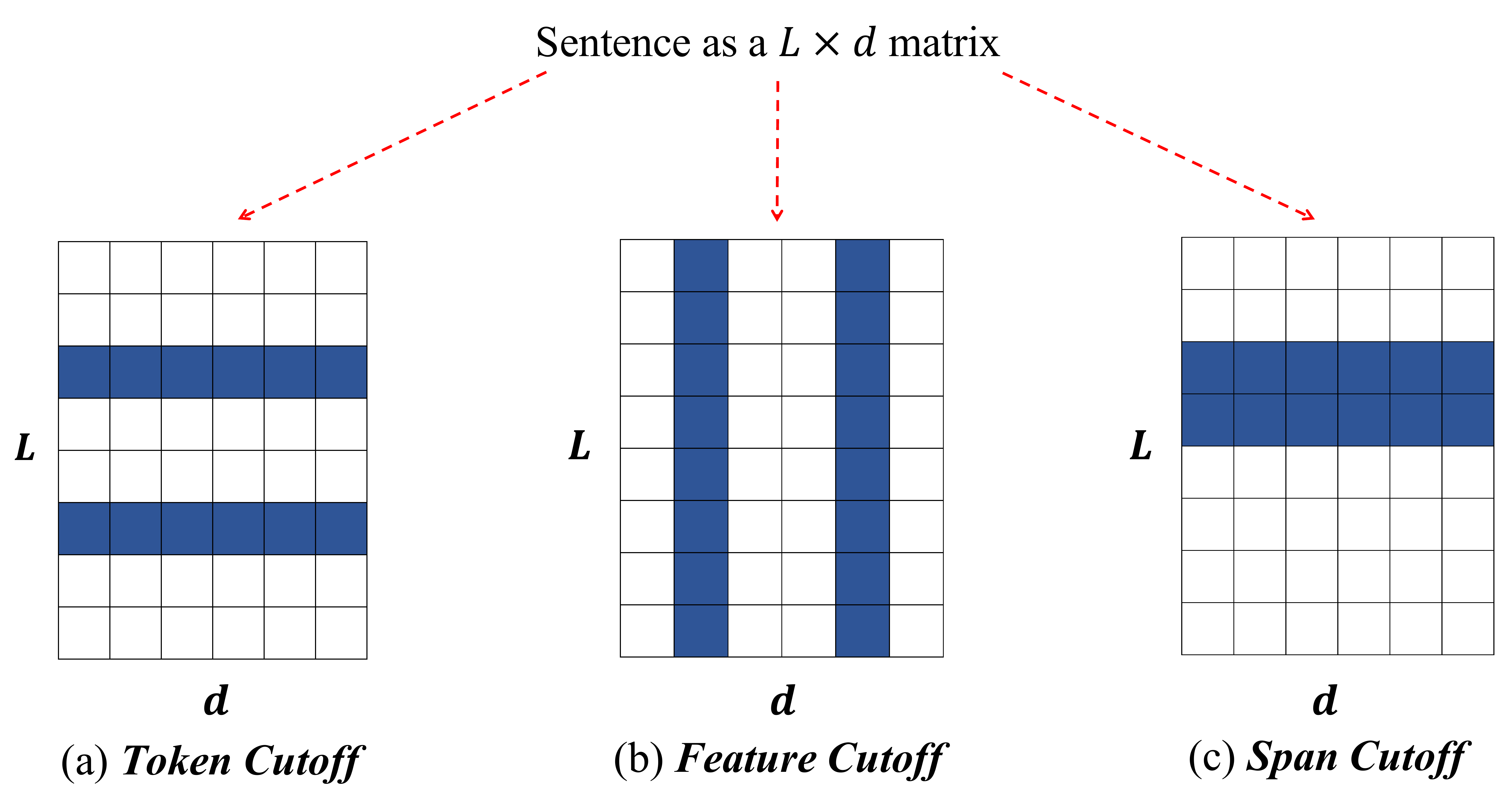} 
	\vspace{-4mm}
	\caption{Schematic illustration of the proposed cutoff augmentation strategies, including token cutoff, feature cutoff and span cutoff, respectively. Blue area indicates that the corresponding elements within the sentence's input embedding matrix are removed and converted to $0$. Notably, this is distinct from Dropout, which randomly transforms elements to $0$ (without considering any underlying structure of the matrix). }
	\label{fig:illustration}
	\vspace{-4mm}
\end{figure*}
\vspace{-4mm}
\subsection{Constructing Partial Views}
\vspace{-1mm}
To obtain partial views of a given sentence, \cite{Clark2018SemiSupervisedSM} proposed to carefully select hidden representations at the top of a Bi-LSTM sentence encoder. However, this strategy is not generic enough since it relies on the unidirectional nature of LSTM. For transformer-based architectures that are widely adopted nowadays, each output hidden unit has access to the information of all the input tokens (given the property of self-attention networks). In this regard, we argue that collecting restricted views at the input embedding space could be a more model-agnostic solution.

Given a text sequence $\mathbf{x} = [x_1, ..., x_L]$, whose input embedding matrix is denoted by $\mathbf{W} \in \mathbb{R}^{L \times d}$. Note that $w_{i, j}$ represents the $j$-th dimension of the embedding vector corresponding to the $i$-th token, and $d$ is the dimension of the input embeddings. 
We suppose that partial views may be obtained by cutting vectors along either dimensions off, hence the proposed approach is dubbed \emph{Cutoff}. 
Cutoff removes the information from the input embedding matrix in a more structured manner, as opposed to Dropout, which randomly sets elements within the matrix to $0$.
Specifically, either the entire embedding of an individual word or one embedding dimension of every word within the sequence are converted to a vector of zeros (see Figure~\ref{fig:illustration}). 

In the context of pre-trained transformer models, such as BERT or RoBERTa, the input embedding matrix consists of tokens, segments and positional embeddings. To make sure that no information corresponding to the removed tokens is left, all three types of embeddings, in the case of \emph{token cutoff}, are converted to $0$. Moreover, all embedding types are considered while sampling the feature dimensions to be erased.  
Intuitively, with augmented data, the learned model is encouraged to be robust enough so that it can produce consistent predictions with a few words removed from the sentence. 
For \emph{feature cutoff}, since each input embedding dimension contains certain semantic information, the model is impelled to encapsulate rich and meaningful features \emph{w.r.t.} each word given that it needs to make the correct predictions with a certain number of features erased entirely.
\vspace{-2mm}
\paragraph{Span Cutoff} Moreover, \cite{joshi2019spanbert} advocated that predicting spans, relative to predicting individual tokens, provides a more challenging objective for self supervision tasks. Thus, we conjecture that easing a contiguous span of text may also lead to harder augmented examples, which can benefit the model during the fine-tuning stage to a larger extent. Therefore, we propose an additional strategy to obtain partial views of the input. First, a preset coefficient $\alpha$ is defined, which indicates the ratio between the length of removed span to that of the original sequence.
Then, to obtain a span with the length of $l = \lfloor \alpha \times L \rfloor$ ($\lfloor \cdot \rfloor$ denotes the floor function), the starting index $s$ for the span is first randomly sampled as: $s \in \{ 0, 1, ..., L - l\}$. Afterwards, the embeddings \emph{w.r.t.} the tokens between the $s$-th and $(s+l-1)$-th positions are all converted to vectors of zeros.

\begin{figure*}
	\centering
	\includegraphics[width=1.0\textwidth]{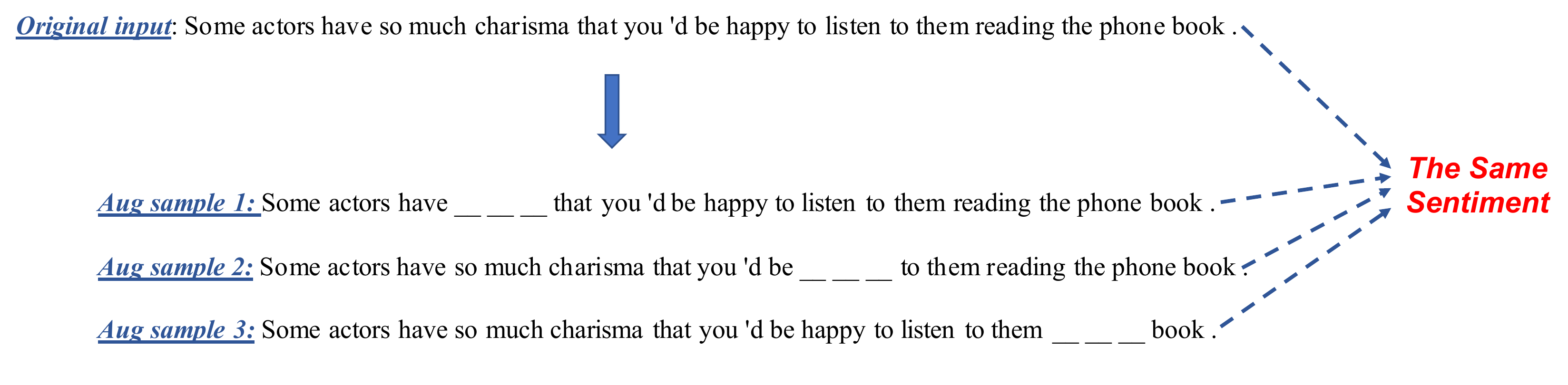} 
	\vspace{-6mm}
	\caption{Illustration of the proposed span cutoff method with one specific example (from the SST-2 dataset). In this case, the model is supposed to produce consistent predictions (\emph{i.e.}, sentiments) for all three augmented samples (with various spans of tokens removed).}
	\label{fig:sentiment}
	\vspace{-4mm}
\end{figure*}

As illustrated in Figure~\ref{fig:sentiment}, \emph{span cutoff} removes a continuous chunk of texts away, and the remaining sentences preserve the same label (\emph{e.g.} sentiment) as the original example. 
Since certain semantic information within a sentence has been removed, these augmented samples could better encourage the model to \emph{fully} leverage different features that may be helpful for predicting the sentiment of the original input (rather than merely relying on a small set of salient ones).

\vspace{-1mm}
\subsection{Incorporating Augmented Samples}
Suppose there are $N$ cutoff samples constructed from the same original input $x$ (with a label of $y$), which are denoted by $x_{\text{cutoff}}^{1}, x_{\text{cutoff}}^{2}, ... ,x_{\text{cutoff}}^{N}$, respectively. Since their semantic meanings are approximately preserved with the cutoff operation, we may incorporate them into the training objective by encouraging the model to make similar predictions across different samples. The training objective can be written as:
\begin{align}
\mathcal{L}= & \ {\mathcal{L}_{\rm ce}(x, y)} +  \alpha \sum_{i=1}^{N} {\mathcal{L}_{\rm ce}(x_{\text{cutoff}}^{i}, y)} \nonumber \\
& +\ \beta\mathcal{L}_{\rm divergence}(x, x_\text{cutoff}^{1}, x_\text{cutoff}^{2}, ..., x_\text{cutoff}^{N}, y) \,,  \label{eq:loss}
\end{align}

\noindent where ${\mathcal{L}_{\rm ce}}$ denotes the cross-entropy loss, which are applied to both original and augmented samples. Furthermore, to explicitly minimize the gap between the predictions \emph{w.r.t} all the sentences, $\mathcal{L}_{\rm divergence}$ is utilized to measure the consensus between all the predictions. KL-divergence has been widely adopted as the divergence metric in previous works \cite{Miyato2017AdversarialTM, miyato2018virtual, Clark2018SemiSupervisedSM, xie2019unsupervised}. However, since there are multiple cutoff samples' predictions, calculating KL divergence in a pair-wise manner will lead to  $2^{N+1}$ terms, and thus can be quite computationally intensive. To this end, we propose to leverage the Jensen-Shannon (JS) divergence consistency loss as $\mathcal{L}_{\rm divergence}$. 
Concretely, the $\mathcal{L}_{\rm divergence}$ term can be obtained as follows:
\begin{align}
p_{avg} & = \frac{1}{N+1} \sum_{i=0}^{N} p(y|x_\text{cutoff}^{i})   \nonumber \\
\mathcal{L}_{\rm divergence}  & = \frac{1}{N+1} \sum_{i=0}^{N} \text{KL}[p(y|x_\text{cutoff}^{i}) || p_{avg}]  \,    \label{eq:js}
\end{align}
To be more specific, the average over all the predictions are first calculated, which is then employed to match with each individual predictions. with such a scheme, the consensus between multiple augmented data along with the original sample is measured in an efficient way. Moreover, it has been shown that the JS divergence loss can endow the model with more stability and consistency across a diverse set of inputs \cite{Bachman2014LearningWP, Zheng2016ImprovingTR, Kannan2018AdversarialLP, Hendrycks2020AugMixAS}. 
\vspace{-0.5mm}
\subsection{Extension to Language Generation}
\vspace{-0.5mm}
The various types of \emph{cutoff} strategies proposed above can be naturally extended to conditional text generation scenario as well. Given a sentence pair $(x_{input}, x_{output})$, the \emph{cutoff} operation can be performed on both sentences to synthesize an augmented training pair. Intuitively, the model has access to a restricted view of the input, while being asked to predict part of the output sequence. It has been shown that neural text generation systems are highly sensitive to input noise \cite{lee2018hallucinations}, and we suppose that adding such augmented examples could improve the model's generalization ability. We evaluate this hypothesis on the machine translation task empirically (see Sec~\ref{sec:mt}).
\subsection{Computational Complexity}
\label{sec:computation}
We now compare the asymptotic complexity of \emph{cutoff} to adversarial-based approaches. FreeLB \cite{zhu2019freelb} and SMART \cite{jiang2019smart} are two representative adversarial training methods applied to NLP domain. Both require additional ascent steps to determine the perturbation directions. Let $T$ denote the number of ascent steps needed for adversarial training, where we have $T \geq 1$. The numbers of forward and backward passes for FreeLB and SMART are both $1+T$. On the other hand, \emph{cutoff} requires no extra backward passes (and thus has a backward pass number of $1$). As to the forward pass, given that the size of augmented samples is the same as that of original training instances, the number of forward passes is doubled to $2$, which is smaller relative to adversarial training. Overall, the \emph{cutoff} approach takes less computational overhead compared to standard SGD-based training.

\section{Experimental Setup}
\subsection{Datasets}
We evaluate the effectiveness of the proposed \emph{Cutoff} approach on both natural language understanding and generation tasks.
To facilitate comparisons with other baseline methods, we employ the GLUE benchmark, which consists of a wide variety of natural language understanding tasks: natural language inference (MNLI, QNLI), textual entailment (RTE), paraphrase identification (MRPC, QQP), sentiment analysis (SST), textual similarity (STS) and linguistic acceptability (CoLA). In terms of the evaluation metrics, accuracy is used for most of the datasets, except for STS and CoLA, where Spearman correlation and Matthews correlation are utilized, respectively. We use Roberta \cite{liu2019roberta} as the testbed for our data augmentation strategies, including both the \emph{base} and \emph{large} models.
Besides the natural language understanding benchmark, we further evaluate the effectiveness of proposed cutoff approach on the neural machine translation (NMT) task. Specifically, WMT14 German-to-English and English-to-German datasets are used as the testbeds, where a Transformer-Base model with 6-layer encoder and 6-layer decoder is employed as the baseline.

\begin{table*}[ht!]
	\centering
	\begin{small}
		\vspace{0mm}
		\setlength{\tabcolsep}{6pt}
		\def\arraystretch{1.05}
		\begin{tabular}{c||c|c|c|c|c|c|c|c|c}
			\toprule[1.2pt]
			\tf{Model} & \tf{MNLI} & \tf{QNLI}    & \tf{QQP} & \tf{RTE}  & \tf{SST-2} & \tf{MRPC}  & \tf{CoLA} & \tf{STS-B} & \tf{Avg} \\
			\hline
			\multicolumn{9}{c}{\emph{Testbed}: \emph{\textbf{RoBERTa-base}}} \\
			\hline
			RoBERTa-base  & 87.6 & 92.8  & 91.9  & 78.7  & 94.8  & -/90.2 & 63.6 & 91.2 & - \\
			ALUM & 88.1 & 93.1  & \textbf{92.0}  & 80.2  & 95.3  & 90.9/- & 63.6 & 91.1 & 86.8 \\
			\hline
			 Token Cutoff & 88.2 & 93.3  &91.9  &  81.2 & 95.1 & \textbf{91.1}/-  & 64.1 & 91.2 & 87.0 \\
			 Feature Cutoff & 88.2 & 93.2  & \textbf{92.0} &  81.6 & 95.3 & 90.7/- & 63.6 & 91.2 & 87.0 \\
			 Span Cutoff & \textbf{88.4} & \textbf{93.4}  & \textbf{92.0}/-  &  \textbf{82.3} & \textbf{95.4} & \textbf{91.1} & \textbf{64.7} & 91.2 & \textbf{87.3} \\
			\hline 
			\multicolumn{9}{c}{\emph{Testbed}: \emph{\textbf{RoBERTa-large}}} \\
			\hline 
			RoBERTa-large & 90.2 & 94.7  & 92.2  & 86.6 & 96.4 & -/90.9 & 68.0 & 92.4 & - \\
			PGD  & 90.5 & 94.9  & 92.5  & 87.4 & 96.4 & 90.9/- & 69.7 & 92.4 & 89.3 \\
			FreeAT  & 90.0 & 94.7  & 92.5  & 86.7 & 96.1 & 90.7/- & 68.8 & 92.4 & 89.0  \\
			FreeLB   & 90.6 & 95.0  & \textbf{92.6}  & 88.1 & 96.8 & \textbf{91.4}/- & 71.1 & 92.7 & 89.8 \\
			ALUM & 90.9 & 95.1  & 92.2  & 87.3 & 96.6 & 91.1/- & 68.2 & 92.1 & 89.2  \\
			SMART  & 91.1 & \textbf{95.6}  & 92.4  & \textbf{92.0} & 96.9 & 89.2/92.1 & 70.6 & 92.8 & 90.0 \\
			Back-translation  & 91.1 & 95.3  & 92.0  & 91.7 & 97.1 & 90.9/93.5 & 69.4 & 92.8 & 90.1  \\
			\hline
			Token Cutoff & 91.0 & 95.3  & 92.3  & 90.6 & 96.9 & 90.9/93.2 & 70.0 & 92.5 & 90.0 \\
			Feature Cutoff  & 90.9 & 95.2  & 92.4  & 90.9 & \textbf{97.1} & 90.9/93.4 & 71.1 & 92.4 & 90.1 \\
			Span Cutoff  &  \textbf{91.1} &  95.3  & 92.4 & 91.0 &  96.9 & \textbf{91.4}/\textbf{93.8} & \textbf{71.5} & \textbf{92.8} & \textbf{90.3} \\
			\bottomrule[1.2pt]
		\end{tabular}%
		\caption{The empirical results on the dev sets of the GLUE benchmark, with both \emph{Roberta-base} and \emph{Roberta-large} models as the corresponding baselines. Notably, the proposed Cutoff strategies deliver competitive numbers while being more computationally efficient.}
		\label{tab:eval}
	\end{small}
	\vspace{-2mm}
\end{table*}

\subsection{Training Details}
We finetune the pre-trained models using Adam \cite{kingma2014adam}, with the learning rate selected from $\{$5e-6, 8e-6, 1e-5, 2e-5$\}$ for all parameters. The same learning rate decay scheme as \cite{liu2019roberta} is employed, with a warmup ratio of 0.06 and a linear decay schedule. We also apply a weight decay  of $0.1$ during training. The max number of epochs is set as either $5$ or $10$. The batch size is chosen as $16$ for all model variants. The coefficients $\alpha$ (corresponding to the cross-entropy loss on the cutoff samples) and $\beta$ (associated with the $\mathcal{L}_{\rm divergence}$ term) are both selected from $\{$0.1, 0.3, 1, 3$\}$ on the validation set.

\subsection{Baselines}
We consider several strong baselines to compare with the proposed methods, which can be approximately divided into two categories: \emph{\romannumeral1}) approaches based on adversarial training, including PGD \cite{Madry2018TowardsDL},  FreeAT \cite{shafahi2019adversarial}, FreeLB \cite{zhu2019freelb}, ALUM \cite{liu2020adversarial}. Notably, these methods are more computationally intensive than \emph{Cutoff}; \emph{\romannumeral2}) other data augmentation strategies for natural language. Back translation is evaluated and compared with our methods given its wide adoption. Consistency training objective is utilized for back translation in our implementation to ensure fair comparison. Although back translation, similar to \emph{Cutoff}, serves as a label-preserving transformation on original training instances, it requires additional data (\emph{i.e.}, language pairs) and translation model pre-training. From this perspective, the \emph{Cutoff} approach is easier to use as a drop-in replacement to standard training.

\section{Experimental Results}
We experimented three different \emph{Cutoff} variants in terms of the strategy to construct partial views, \emph{i.e.}, token cutoff, feature cutoff and span cutoff. They are evaluated and compared on the GLUE benchmark. Detailed analysis and ablation studies regarding the cutoff approach are further conducted, where the advantage of utilizing the JS divergence framework is demonstrated. Besides, we also investigate the effectiveness of the \emph{Cutoff} approach on German-to-English and English-to-German machine translation tasks. 

\subsection{GLUE Benchmark Evaluation}
The empirical results of proposed \emph{Cutoff} strategies (relative to other strong baselines) are presented in Table~\ref{tab:eval}. It can be observed that the different \emph{Cutoff} methods consistently outperform ALUM on top of the RoBERTa-\emph{base} model, while being much more computationally efficient (see Section~\ref{sec:computation}). Moreover, span cutoff delivers the strongest numbers on most datasets, which aligns with our assumption that easing a span from the input sequence could lead to more challenging and thus more useful augmented samples. 

As to the case where RoBERTa-\emph{large} is employed as the baseline, the \emph{cutoff} data augmentation strategies again consistently exhibit competitive or better performance compared with several adversarial-based approaches. 
It is worth noting that the Cutoff approaches are related to adversarial-based training in the sense that they both try to produce additional samples with certain perturbations around the original input. However, adversarial-based methods require additional computations to determine the perturbation directions, whereas Cutoff simply remove one slice of information from  the input embedding matrix (which could be at the token, feature or span level). This leverages the prior knowledge that the information is organized in a structured manner within the input embeddings, and thus a model with strong generalization ability should be able to make consistent predictions while only partial views are available.

Moreover, compared with back translation, the \emph{Cutoff} approaches also demonstrate the same or stronger results on $6$ out of $8$ NLU tasks considered here. This further verifies the effectiveness of \emph{Cutoff} as a data augmentation strategy despite its simplicity.  

\subsection{Application to Machine Translation}
\label{sec:mt}
To investigate the effectiveness of \emph{cutoff} on text generation problems, we further apply it to the neural machine translation tasks. Specifically, we leverage the $6$-layer Transformer Base architecture \cite{vaswani2017attention} as the baseline. Cutoff is applied to both the input and output sequences to produce their partial views, which are used as augmented translation pairs for training purpose. 
To ensure fair comparison, the same beam decoding configuration with \cite{vaswani2017attention} is utilized. 

\begin{table}[ht!]
	\centering
	\begin{small}
		\vspace{0mm}
		\setlength{\tabcolsep}{6pt}
		\def\arraystretch{1.18}
		\begin{tabular}{c||c}
			\toprule[1.2pt]
			\tf{Model} & \tf{BLEU score} \\
			\hline
			Transformer Base \cite{vaswani2017attention} & 27.3 \\
			Admin \cite{Liu2020UnderstandingTD} & 27.9 \\ 
			Transformer Base\tablefootnote{This number is reported in \cite{So2019TheET} for the Transformer Base model. The same evaluation settings are used for our cutoff method, \emph{i.e.}, case-sensitive tokenization and the compound splitting are both used.} \cite{So2019TheET}  & 28.2 \\
			Evolved Transformer \cite{So2019TheET} & 28.4 \\
			Weighted Transformer \cite{Ahmed2017WeightedTN} & 28.4 \\
			Adversarial Training \cite{Wang2019ImprovingNL} \ & 28.4 \\
			\hline
 			Transformer Base \& Cutoff (w/o  JS loss) & 28.9 \\
    		Transformer Base \& Cutoff (w/  JS loss) & \textbf{29.1} \\
			\bottomrule[1.2pt]
		\end{tabular}%
		\caption{BLEU scores of the proposed cutoff method on the WMT2014 English-to-German machine translation task, compared with adversarial-based baselines. All methods are built on top of $6$-layer Transformer Base model \cite{vaswani2017attention}.}
		\label{tab:mt_eval}
	\end{small}
	\vspace{-2mm}
\end{table}
In the initial experiments, we found that \emph{token cutoff} performs the best on machine translation tasks. This may be attributed to the fact that removing spans from both the source and target sentences would result in large information mismatch between the input and output, and thus the resulting pairs may be too challenging. The results on the WMT2014 English-to-German dataset are presented in Table~\ref{tab:mt_eval}. Relative to several competitive baseline methods that are based upon $6$-layer Transformer Base model, our token cutoff approach exhibits the best BLEU score. More importantly, cutoff outperforms the adversarial training approach introduced by \cite{Wang2019ImprovingNL}. Concretely, they proposed to inject adversarial perturbations on the output word embeddings (in the softmax layer). Notably, their adversarial training strategy requires updating the model parameters and adversarial perturbation alternately. Thus it is more complicated and computationally expensive than our approach. Besides, it is observed that the JS divergence objective leads to further gains, demonstrating its complementary nature with the standard cross-entropy objective.
\begin{table}[ht!]
	\centering
	\begin{small}
		\vspace{0mm}
		\setlength{\tabcolsep}{6pt}
		\def\arraystretch{1.18}
		\begin{tabular}{c||c}
			\toprule[1.2pt]
			\tf{Model} & \tf{BLEU score} \\
			\hline
			Actor-critic \cite{bahdanau2016actor} & 28.5 \\
			Transformer Base \cite{vaswani2017attention} & 34.4 \\
			Adversarial training \cite{Wang2019ImprovingNL} & 35.2 \\
			Data Diversification \cite{Nguyen2019DataDA} & 37.2 \\
			MAT \cite{fan2020multi} & 36.2 \\
			Mixed Representations \cite{wu2020sequence} & 36.4 \\
			MAT+Knee \cite{iyer2020wide} & 36.6 \\
			\hline
			Transformer Base \& Cutoff (w/o  JS loss) & 36.7 \\
    		Transformer Base \& Cutoff (w/  JS loss) & \textbf{37.6} \\
			\bottomrule[1.2pt]
		\end{tabular}%
		\caption{BLEU scores of the proposed cutoff method on the IWSLT2014 German-to-English machine translation task, relative to adversarial-based baseline and other state-of-the-art models.}
		\label{tab:mt_eval_2}
	\end{small}
	\vspace{-2mm}
\end{table}
In addition, on the IWSLT2014 German-to-English dataset, \emph{Cutoff} again consistently exhibits significant gains over the Transformer Base model. Along with the JS loss term introduced, our model achieves a BLEU score of $37.6$, greatly outperforming the adversarial-based method. As shown in Table~\ref{tab:mt_eval_2}, by simply employing \emph{Cutoff} on top of a $6$-layer Transformer model, our approach outperforms all the previous works, further demonstrating its effectiveness.
\vspace{-1mm}
\subsection{Ablation Study}
\vspace{-1mm}
\subsubsection{The effect of JS divergence loss}
To investigate the importance of incorporating the Jensen-Shannon (JS) divergence consistency loss, we select different values of $\beta$ (ranging from 0.0 to 3.0) and explore how the dev set results (on the MNLI dataset) would change accordingly. The coefficient \emph{w.r.t.} the cross-entropy (CE) loss term is set as $1$ for all the ablation settings, and $\beta$ controls the relative weight of the (JS) divergence consistency loss term. As shown in Table~\ref{tab:beta}, leveraging the JS loss term consistently improves the empirical performance (relative to only using the CE loss term), and a $\beta$ value of $1$ gives rise to the best empirical result on the MNLI dataset.
\begin{table} [ht!]
	\centering
	\def\arraystretch{1.25}
	\footnotesize 
	\vspace{-1mm}
	\begin{tabular}{c||c|c|c|c|c}
		\toprule[1.2pt]
	    $\boldsymbol{\beta}$ & $0.0$ & $0.1$ & $0.3$ & $1.0$ & $3.0$  \\
		\hline
		\textbf{Accuracy} & 88.21 & 88.27  & 88.32  & $\tf{88.36}$  & 88.12  \\
		\bottomrule[1.2pt]
	\end{tabular}
	\caption{Ablation study for the span cutoff augmentation strategy with different choices of $\beta$, the coefficient \emph{w.r.t.} the JS divergence loss term (the cross-entropy coefficient $\alpha$ is set as $1$). The performance is measured with accuracy on the dev set of the MNLI dataset.}
	\label{tab:beta}
	\vspace{-2mm}
\end{table}
\vspace{-2mm}
\subsubsection{The effect of \emph{Cutoff} ratios}
\vspace{-1mm}
One important hyperparameter with the \emph{Cutoff} approach is the ratio of elements to be removed, where the \emph{elements} can be tokens, features or a span (depending on the specific Cutoff variant). The ratio can be regarded as the magnitude of perturbations applied to the input sentence.

\begin{figure} [ht!]
	\centering
	\includegraphics[width=0.48\textwidth]{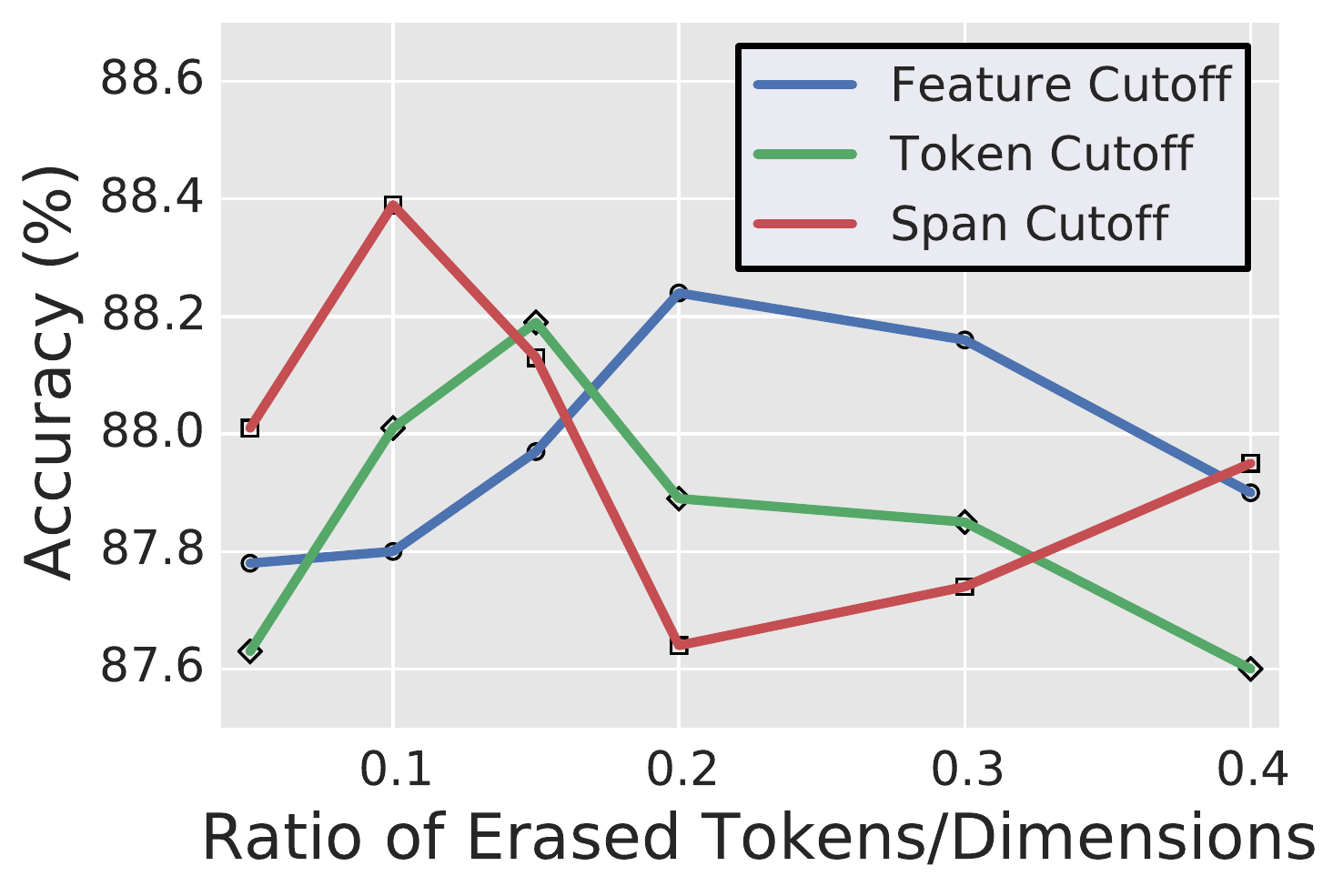} 
	\vspace{-2mm}
	\caption{The MNLI dev set performance with different Cutoff strategies across different choices of cutoff ratios, which refers to the ratios corresponding to the number of removed tokens, the number of removed feature dimensions or the erased span length, respectively.}
	\label{fig:cutoff_ratios}
	\vspace{-2mm}
\end{figure}

As shown in Figure~\ref{fig:cutoff_ratios}, we applied various cutoff ratios to the different Cutoff variants, including $0.05$, $0.1$, $0.15$, $0.2$, $0.3$ and $0.4$. It can be observed that determining a sweet point of the ratio is critical to the generalization ability of the resulting model. Specifically, token cutoff shows the best performance with a ratio of $0.15$, whereas feature cutoff gives rise to the strongest number at a ratio of $0.2$. Span cutoff, on the other hand, performs the best with a ratio of $0.1$ (the length of the removed span \emph{w.r.t} the entire sentence). Using a ratio that is too large tends to result in smaller improvements. This may be attributed to the fact that the assumption that the label of the original data is preserved does not hold true (with larger perturbations).


\section{Conclusion}
\vspace{-2mm}
In this paper, we introduced \emph{cutoff}, a set of data augmentation strategies that can serve as a drop-in replacement to enrich original training data. The augmented samples are produced stochastically by obtaining partial views of an input sentence. Notably, this process requires no additional computational overhead, and is thus more efficient than adversarial-training-based approaches (which involves additional backward operation to determine the perturbation directions). With extensive experiments on natural language understanding and machine translation tasks, \emph{cutoff} gave rise to significant gains, and performed on par or stronger than several competitive baselines based upon adversarial training (while taking a fraction of training time). It is worth noting that \emph{cutoff}, combined with the proposed JS divergence loss, achieved state-of-the-art result on the IWSLT2014 German-English dataset, with a test BLEU score of $37.6$.

\section*{Acknowledgements}
We thank Sandra Sajeev for her help to edit and refine the paper. 

\bibliography{acl2020}
\bibliographystyle{acl_natbib}


\end{document}